\documentclass[english, 12pt, a4paper]{article}
\usepackage[T1]{fontenc}
\usepackage[latin9]{inputenc}
\usepackage{babel}
\usepackage{array}
\usepackage{booktabs}
\usepackage{amsmath}
\usepackage{amsthm}
\usepackage{amsfonts}
\usepackage{graphicx}
\usepackage[figurename=Figure]{caption}
\usepackage{subfig}
\usepackage{float}
\usepackage{authblk}
\usepackage{wrapfig}
\usepackage[unicode=true,pdfborder={0 0 1},backref=false,colorlinks=false]
{hyperref}
\usepackage{adjustbox}
\usepackage{siunitx}

\makeatletter
\setkeys{Gin}{width=\linewidth}
\renewcommand{\[}{\begin{equation}}
\renewcommand{\]}{\end{equation}}
\providecommand{\doi}[1]{%
	\begingroup
	\let\bibinfo\@secondoftwo
	\urlstyle{rm}%
	\href{http://dx.doi.org/#1}{%
		doi:\discretionary{}{}{}%
		\nolinkurl{#1}%
	}%
	\endgroup
}

\@ifundefined{showcaptionsetup}{}{%
 \PassOptionsToPackage{caption=false}{subfig}}

\providecommand{\keywords}[1]{\textbf{\textit{Keywords:\textbf{}}} #1}

\makeatother

\begin{document}

\title{Binary Stochastic Filtering: a Method for Neural Network Size Minimization and
Supervised Feature Selection}

\author[a]{A.~Trelin}
\author[b]{A.~Procházka}

\affil[a]{Department of Solid State Engineering, University of Chemistry and
	Technology, 16628 Prague, 
	Czech Republic\newline \href{mailto:trelina@vscht.cz}{trelina@vscht.cz}}

\affil[b]{Department of Computing and Control Engineering, University of Chemistry and 
	Technology, 16628 Prague, Czech Republic \newline \href{mailto:a.prochazka@ieee.org }{a.prochazka@ieee.org }}

\maketitle
\thispagestyle{empty}

\begin{abstract}
Binary Stochastic Filtering (BSF), the algorithm for feature selection
and neuron pruning is proposed in this work. The method defines 
filtering layer which penalizes amount of the information involved in the
training process.
This information could be the input data or output of the previous layer,
which directly lead to the feature selection or neuron pruning respectively,
producing \textit{ad hoc} subset of features or selecting optimal
number of neurons in each layer.

Filtering layer stochastically passes or drops features 
based on individual weights, which are tuned with standard backpropagation 
algorithm during the training process.
Multifold decrease of neural network size has been achieved in the experiments. 
Besides, the method was able to select minimal number of features, 
surpassing literature references by the accuracy/dimensionality ratio.

Ready-to-use implementation of the BSF layer is provided.
\end{abstract}

\clearpage

\pagenumbering{arabic} 
\section{Introduction}

Supervised learning algorithms rely on prearranged labeled datasets
as a source of information.
In a variety of cases a
dataset consists of a significant number of various features (e.g.
versatile statistical measures for text or network data
processing), with only some of them containing sensible information 
for solving given problem (usually classification or regression).
The process of finding an information-rich subset is called feature
selection and is of growing interest during modern development
of machine learning. Correct feature selection aims to reduce the computational
complexity, to enhance generalization \cite{bermingham2015application}
or may be of interest on its own for further interpretation.

\begin{figure}[h]
	\centering
	\includegraphics[width=0.55\textwidth]{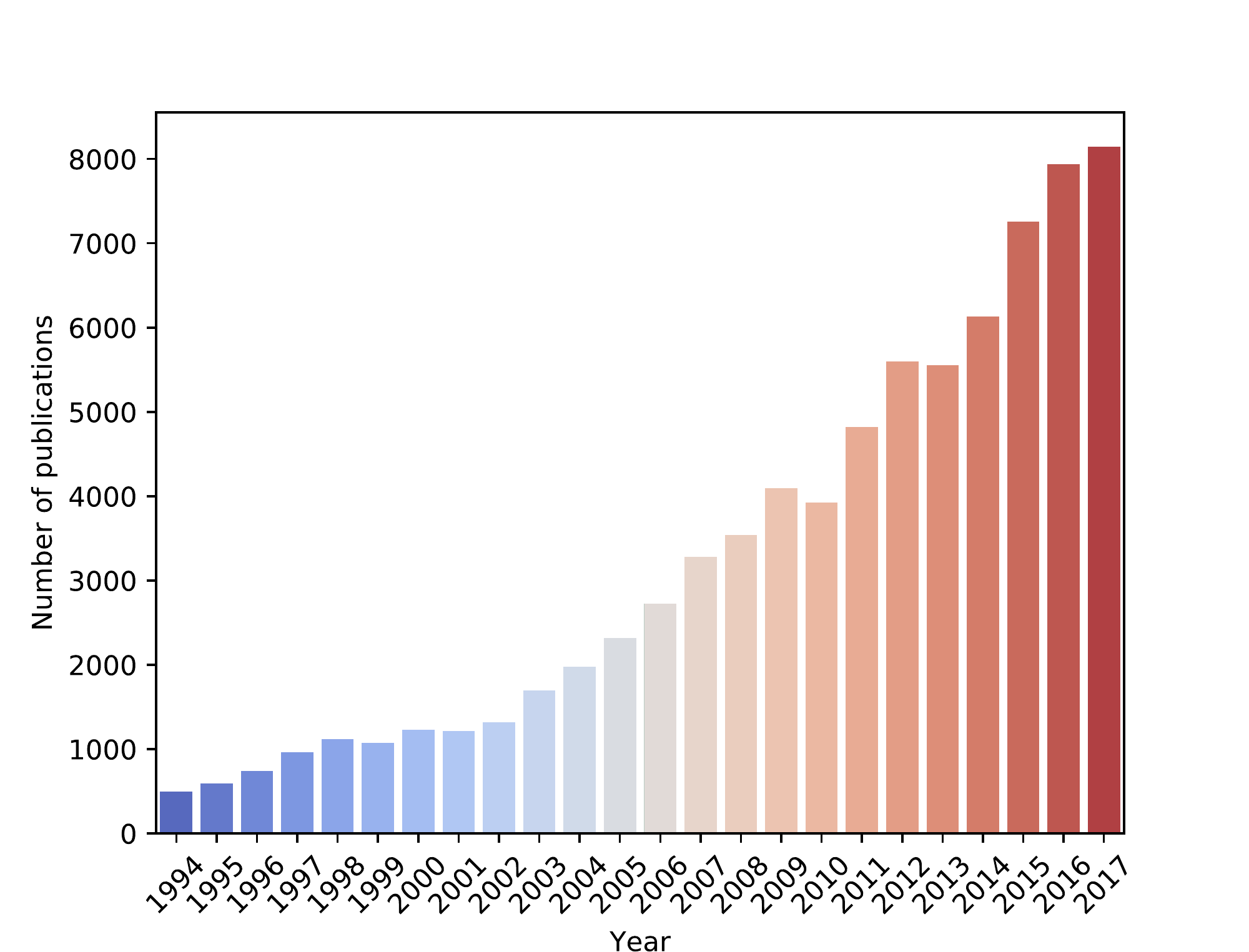}
	\caption{\label{fig:Number-of-publications}Number of publications considering
		feature selection in past years. Image is generated based on data
		from the Web of Science service \cite{reuters2011isi}, searched
		for topic: ``feature selection''. }
\end{figure}

Numerous approaches were proposed in the last 25 years (Fig. \ref{fig:Number-of-publications}),
which include algorithms based on information theory, that tries
to estimate the amount of information, brought to the dataset by a
given feature \cite{fleuret2004fast,peng2005feature,yu2003feature},
Relief and derived algorithms \cite{kira1992practical,robnik2003theoretical},
based on nearest neighbor search, statistical \cite{liu1995chi2}
and spectral graph theory based \cite{zhao2007spectral}, etc. The
very modern approaches are mainly focused on the application of swarm
intelligence algorithms \cite{gu2018feature,mafarja2018evolutionary,hancer2018pareto}
Nevertheless, different methods usually lead to strongly
discrepant sets of features and final decision could be made only
after a series of experiments.

This paper proposes \textit{in situ} feature selection method which relies on the intrinsic
optimization algorithms, used in machine learning without any external entities. 
Filtering layer performs search for least possible feature involvement by
minimization of feature passing probabilities.
Furthermore, filtering may be utilized for neuron or kernel pruning, 
minimizing the size of a network similarly to features,
which leads to multifold increase of NN performance.
The method can be applied to different kinds of NN, optimized for
various datasets, which is demonstrated in this work.

\section{Method}

Stochastic neurons, the units with nondeterministic behavior are known
for a long time, at least since mid 1980s, when Boltzmann machines
became popular \cite{hinton1984boltzmann}. Besides, the stochastic
neurons are of interest because of their closeness to the biological
ones \cite{chen2010real}. The successful networks, consisting of stochastic
neurons were built \cite{suri2013bio}. Nevertheless, mixing of stochastic
and deterministic units in one network is not common, except
for Dropout technique \cite{srivastava2014dropout}, which works
by randomly disabling deterministic neurons to prevent overfitting.

The proposed algorithm is inspired by both stochastic neurons and dropout.
The method combines two ideas mentioned above and could
be thought as a generalization of dropout with trainable individual
dropout rates,
or as incorporation of the special version of stochastic
neurons into classical deep NN structure. Proposed binary stochastic
filtering (BSF) method works by setting the BSF layer (consisting
of the BSF units) after the layer, outputs of which are to be filtered.
BSF unit stochastically passes the input without changes or sets
it to zero with probabilities based on the unit weights. Strictly speaking,
BSF unit is described as:
\[
BSF(x, w, z)=x_{z<w}
\]
where $w$ represents the adjustable weight of a given unit and $z$ is
uniformly sampled from the range $[0,1]$. 
This definition is inspired by the formula from the work of Bengio et al.
(\cite{bengio2013estimating}, Eq. (3)). Authors have defined the Stochastic
Binary Neuron as $h_i=f(a_i, z_i)=1_{z_i>sigm(a_i)}$, where neuron activation
$a_i$ is given as a weighted sum of its inputs. This definition is directly 
derived from the classical deterministic neurons and could be thought as a noisy modification of them. 
In contrast, BSF includes the single input which passes through the unit
if random value $z$ is smaller than weight $w$ or nullified instead. 
Thus, probability of output to be $x$ is equal to the probability
of $z$ to be less than $w$, $\mathbb{P}(x)=\mathbb{P}(z<w)$, which is equal to
$F_z(w)$, where $F_z$ is a cumulative distribution function for the
variable $z$. From the properties of uniform distribution, 
\[
F_z(w)=\begin{cases}
0, & \text{\ensuremath{\mbox{for} \ w<0}}\\
{w-a}\over{b-a}, & \text{\ensuremath{\mbox{for} \ w \in [a,b)}}\\
1, & \text{\ensuremath{\mbox{for} \ w \geq 0}}
\end{cases}
\]
on the interval $[a,b]$ from which $z$ is sampled, $F_z(w)=w$, as
$[a,b]=[0,1]$. In other words, BSF multiplies the input by 
random variable $b$, sampled from Bernoulli distribution with 
parameter equal to the unit weight $w$:
\[
BSF(x, w) = x \cdot b, ~b \sim Bernoulli(w)\
\]
This allows selective filtering of the layer inputs with
adjustable probabilities of the given input to be filtered, which
equals the weight $w$, supposing $w \in [0,1]$. Setting $w$ to be close
to 0 means that the given feature is close to being completely filtered out, 
while in case of $w$ being close to 1, feature passes through BSF 
without changes. Thereby, by tuning the BSF layer weights it is possible to
control how much each of the features participates in the training
process.

From the obvious reasons, the gradient of a stochastic unit is stochastic
as well, being equal to zero or $\infty$ randomly. Thus, backpropagation
training, being a variation of gradient descent method is not able
to optimize weights of the BSF units. G. E. Hinton has suggested a
straight through (ST) estimator \cite{hinton2012st}, which defines
the gradient of stochastic unit with respect to its input $\frac{\partial BSF(x)}{\partial x}=1$.
Few modifications of this approach were reported, such as multiplication
of the gradient by a logistic sigmoid derivative, i.e. $\frac{\partial BSF(x)}{\partial x}=\frac{{d\sigma(x)}}{d x}$,
where $ \sigma(x) = {1 \over 1+e^{-x}} $
but it was found that simpler unity gradient achieves better performance
\cite{bengio2013estimating} as it does not suffer from vanishing
gradients issue. 
ST estimator is especially suitable
when stochastic neurons are mixed with normal ones, as it does not
distort the gradient between classical units. Proof of the last
statement is given in the Appendix \ref{proof:gradients_independence}.

Thus, by applying the ST estimator, weights of BSF layers could
be optimized during model training.
Moreover, penalizing of the filter weights will lead to a decrease of
above-lying layer involvement in the NN working process, which could be used
for neuron pruning or overfitting prevention. Placing the penalized
BSF layer as the input layer of the NN will decrease probabilities of
feature occurring in correspondence to their importance, which is
equivalent to the feature selection.

\subsection{Feature selection}

\subsubsection{\label{subsec:Deep-NN-classifier}Deep NN classifier }

\begin{figure}[h]
	\centering
	\includegraphics[width=0.45\textwidth]{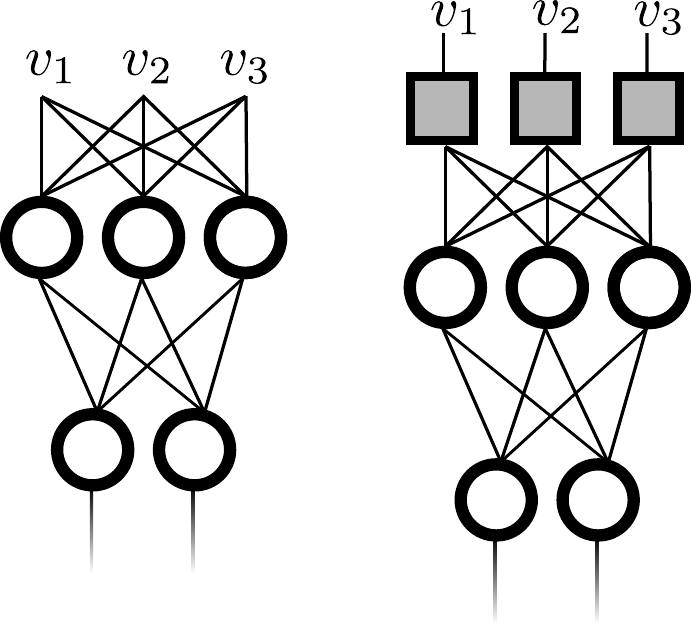}
	\caption{Structure of classical multilayer perceptron (left) and
	same network with included BSF layer. Deterministic and BSF units 
	are drawn in circles and squares respectively, $v_{1..3}$ correspond 
	to input features.}
\end{figure}

Classifier is a model, performing mapping from n-dimensional data
space $D^{n}\rightarrow L$ to the label space consisting of finite (and
usually relatively low) number of labels. As a deep NN classifier,
a classical multilayer perceptron model with softmax activation of the output
layer was used \cite{bridle1990probabilistic}. Softmax function
is a generalization of logistic function for a multilabel classification
problem, defined as 
\[
\sigma_{i}(\vec{x})=\frac{{e^{x_{i}}}}{\sum_{j=1}^{N}{e^{x_{j}}}}
\]
for $i=1,2,...,N$ where $N$ describes number of classes in a given problem. Unlike
classical binary logistic regression, which outputs the scalar
probability $p$ of a data to be true and $1-p$ to be false, softmax
function outputs a vector $\vec{{p}}$ with length $N$, each element
of which represents the probability of the input sample to belong to each
of the classes. Normalization by the sum of outputs guarantees that
total probability of a sample to belong to
all classes will always be unity. Exploitation of the softmax function
needs label to be one hot encoded, i.e. to the vector with length
$N$ 
\[
L_{oh}(i)=\begin{cases}
1, & \text{if \ensuremath{i=classindex}}\\
0, & \text{elsewhere}
\end{cases}
\]

In that case loss function must describe divergence between $\vec{{p}}$
and encoded label $\vec{{L_{oh}}}$, i.e. maps two vectors to a scalar.
Cross-entropy, a function derived from the maximum likelihood estimation method
is widely used as a loss \cite{bishop1995neural}. The function is defined
as 
\[
E=-\sum_{i=1}^{N}{c_{i}\cdot ln(p_{i})+(1-c_{i})\cdot ln(1-p_{i})}
\]
where $c_{i}$ is the actual class label and $p_{i}$ is corresponding
element of $\vec{{p}}$. The selected NN model consists
of 4 layers with rectified linear activation \cite{nair2010rectified}
in the hidden layers. Number of units in each layer was empirically
chosen as $D$, $D$, $2D$, $D$, $N$ respectively, where $D$ is the dimensionality
of input data, in order to meet the different number of features in used datasets.
The model was optimized using Adam algorithm \cite{kingma2014adam} with
parameters $\alpha=0.001$, $\beta_{1}=0.9$, $\beta_{2}=0.999$,
as it is specified in the original paper with zero learning rate decay.
The same optimizer was used for training of all other models.

Classification accuracy for every dataset was estimated using 10-fold
cross-validation method. The NN was trained until full convergence (no loss decrease), both training and validation mean
accuracies were used for evaluation of model performance.

Binary stochastic filter was placed as an input layer to the NN of
the same structure as classifier, resulting in 5 layer network. $L_{1}$
regularization was used as a penalty, with the regularization coefficient
defining the tendency of the data to be filtered out. Generally, this coefficient
serves as a metaparameter in the filtering layer, and needs to be adjusted
to meet the classifier loss. Good results were achieved with regularization
coefficient lying in the range 0.001-0.05. The examples of training
process visualization depending on the regularization rate are provided in
the supplementary materials. The model was trained until convergence.
After model fitting, weights of a BSF layer were analyzed and treated
as a feature importances. The highest importance features
were manually selected and saved forming a reduced dataset. Afterward,
normal classifier was trained as described above for estimation of
classification accuracy for the reduced dataset. 
\label{silhouttes}
To estimate changes in high-dimensional cluster separation, the simple and
well-known silhouettes method \cite{rousseeuw1987silhouettes} was used.
In brief, the method defines silhouette $s(i)$ which compares mean
inter-cluster distance $a(i)$ for the given data point $i$ and mean
distance to all objects belonging to the nearest cluster $b(i)$.
\[
s(i)=\frac{b(i)-a(i)}{max\{a(i),b(i)\}}
\]

The mean value of silhouettes (silhouette coefficient) for all the
datapoints within dataset is a measure for cluster separation, lying
in the range $[-1,1]$ where higher number means better separation.

\begin{figure*}[t]
	
	\centering
	\begin{minipage}[t]{0.4\textwidth}
		\subfloat[]{
			\includegraphics{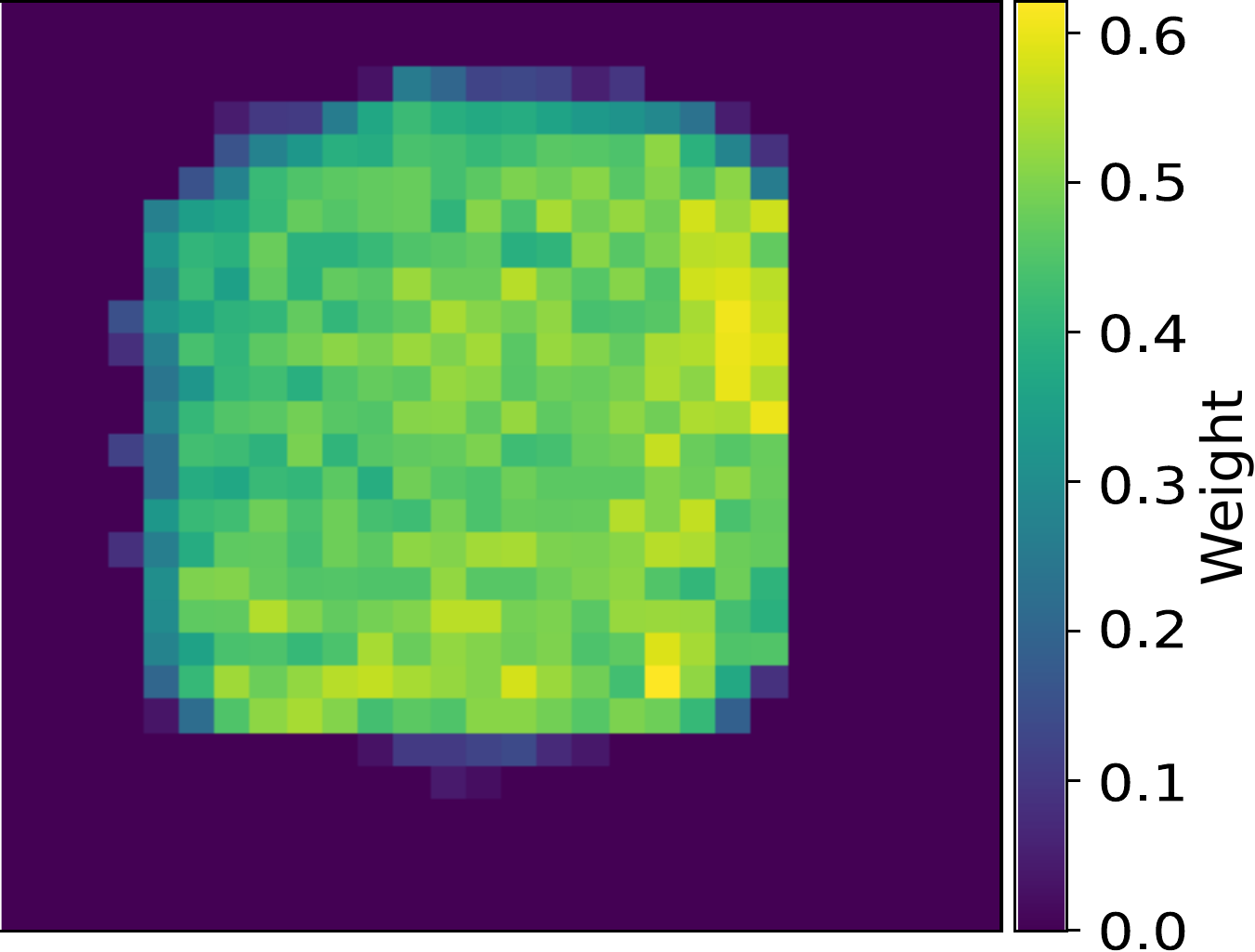}}
	\end{minipage} 
	\vspace{5pt}
	
	\begin{adjustbox}{width=1.0\textwidth,center}
		\begin{minipage}[t]{0.5\textwidth}%
			\subfloat[]{
				\includegraphics{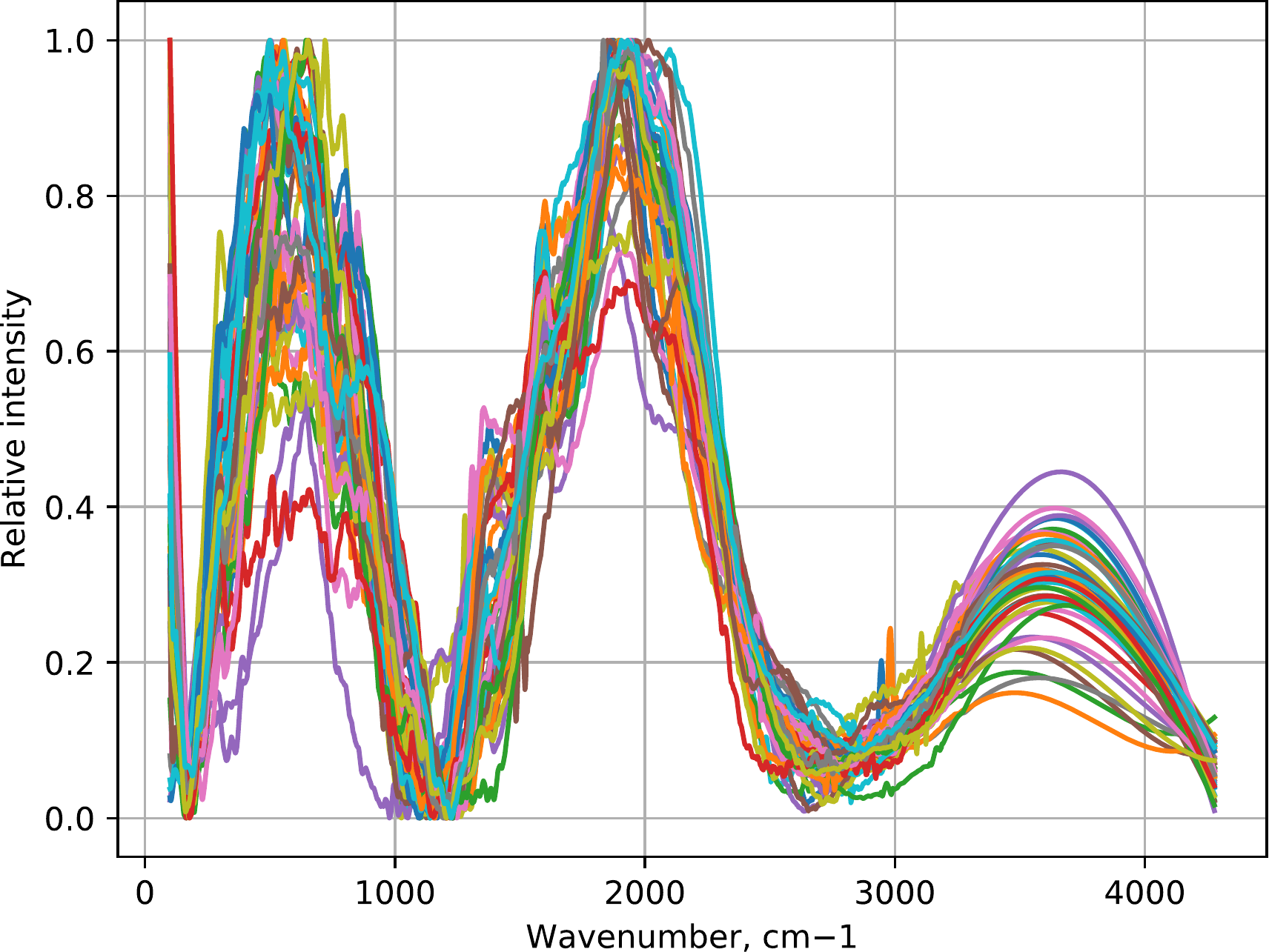}}
		\end{minipage}%
		\hspace{0.2cm}
		
		\begin{minipage}[t]{0.5\textwidth}%
			\subfloat[]{
				\includegraphics{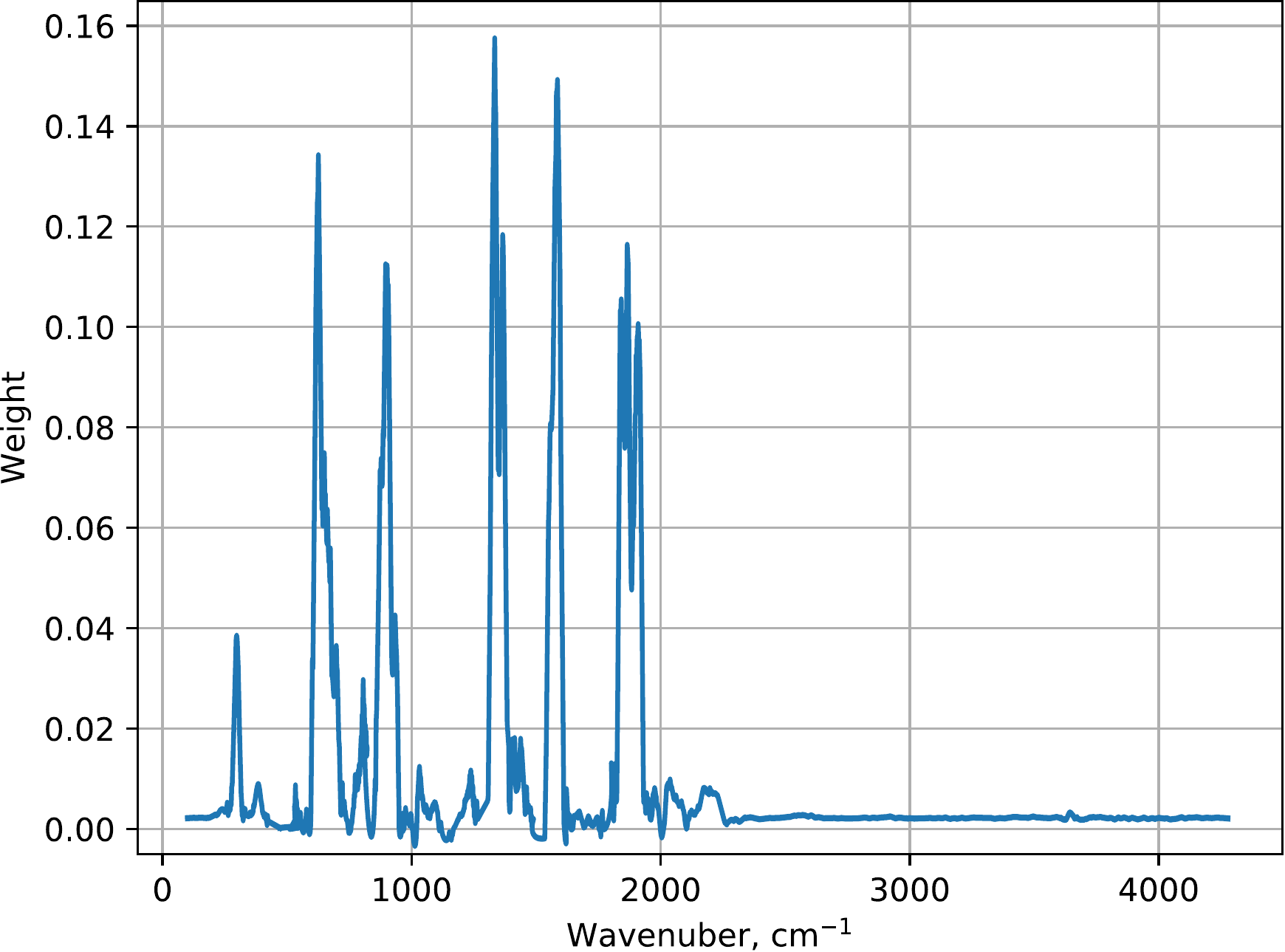}}
		\end{minipage}
	\end{adjustbox}
	\caption{\label{fig:spectra} (a) Weights of the BSF layer after fitting MNIST
		dataset. (b) Set of Raman spectra of different classes after normalization.
		(c) Regions of interest of the spectra selected by BSF.}
	
\end{figure*}

\subsubsection{Convolutional NN classifier}
\label{image_recognition_section}

Convolutional NN (CNN) makes use of data spatial independence, which
is especially useful in image recognition.
The CNN architecture was proposed by Le Cun et al. \cite{lecun1990handwritten}
for number recognition problems. Generally, architecture implements
discrete convolution operation, which converts input data (image) to
the feature maps. Convolution kernel is optimized during the training
process to minimize the loss function value. Produced maps of features are
convoluted further, producing higher-order maps.
After few subsequent convolutional layers the classical fully-connected
classifier (described in the previous part) is placed, which performs actual
classification \cite{lecun1998gradient}. 

The CNN classifier with structure, described in the Table \ref{tab:Convolutional-NN-classifier}, was implemented
in the experiments. 
Its architecture resembles typical simple CNNs, widely used in for educational
and research purposes.
First 3
layers form the convolutional block, which performs feature 
extraction from images, while 3 last layers correspond to
the fully-connected classifier. The model was trained in the same manner
as the DNN classifier described in the previous part (Subsection \ref{subsec:Deep-NN-classifier}). 

Interpretation of feature selection results for image recognition
problem is not that straightforward in comparison to the simple perceptron
classifier, as the convolution operations need input data to
maintain well-defined shape. Nevertheless, its patterns, being visualized,
are easily recognizable. In that case, BSF provides attention maps, 
i.e. highlights the regions
of images with the highest information density, averaged over whole dataset.

\subsubsection{Spectra recognition}

The interpretation of complex compounds spectra is a known problem
in analytical chemistry. This kind of data contains information
about characteristic vibrations of molecules, encoding its structure.
Classically, spectra are manually analyzed by chemists, taking into account
shape and position of spectral peaks. Due to the finite resolution
of the spectroscopes, if a large number of peaks are present 
(i.e. for complex molecules) peaks overlap and interfere, making
spectra impossible to interpret by human.
Few reports on the successful use of machine
learning techniques for identification of the biological samples were
published \cite{GNIADECKA2004443,sigurdsson2004detection}. The binary
stochastic filter was formulated during work on a similar problem,
the classification of DNA spectra. Such spectra are not readable by
human due to presence of thousands of different vibration frequencies. 
Spectra were classified
by the CNN model consisting of 3 convolutional and 4 fully-connected
layers with satisfactory results. After that, binary stochastic filter
layer was added to the model input and training process was repeated.
Weights of the BSF layer, in that case, correspond to the spectral regions
of interest. Weights of the layer were visualized (Fig. \ref{fig:spectra}
(b)) and treated as regions of interest of spectra. Correspondence
between them and expected vibrational frequencies of the DNA constituents
\cite{madzharova2016surface} were observed. This method allows
obtaining information about chemical changes in a set of similar
compounds or mixtures, spectra of which are indistinguishable by other
methods.

\subsection{Neuron pruning}

Sparse neural networks are believed to be more efficient than popular
fully-connected systems due to enhanced generalization, faster
convergence and higher performance \cite{lecun1990optimal}. An alternative
way to sparsifying is neuron pruning, the removal of the least important
units from the network, which was demonstrated by Mozer and Smolensky
\cite{mozer1989skeletonization}. The latter approach seems to be
more beneficial from the performance point of view as removal of the single
unit leads to the greater decrease of computational complexity, nevertheless
the former became more frequent, being popularized by the above-mentioned
work of Le Cun \cite{lecun1990optimal}. Furthermore, the analysis
of units importance is less complex because of their significantly lower amount
in comparison to weights.
\begin{figure}[h]
	\includegraphics{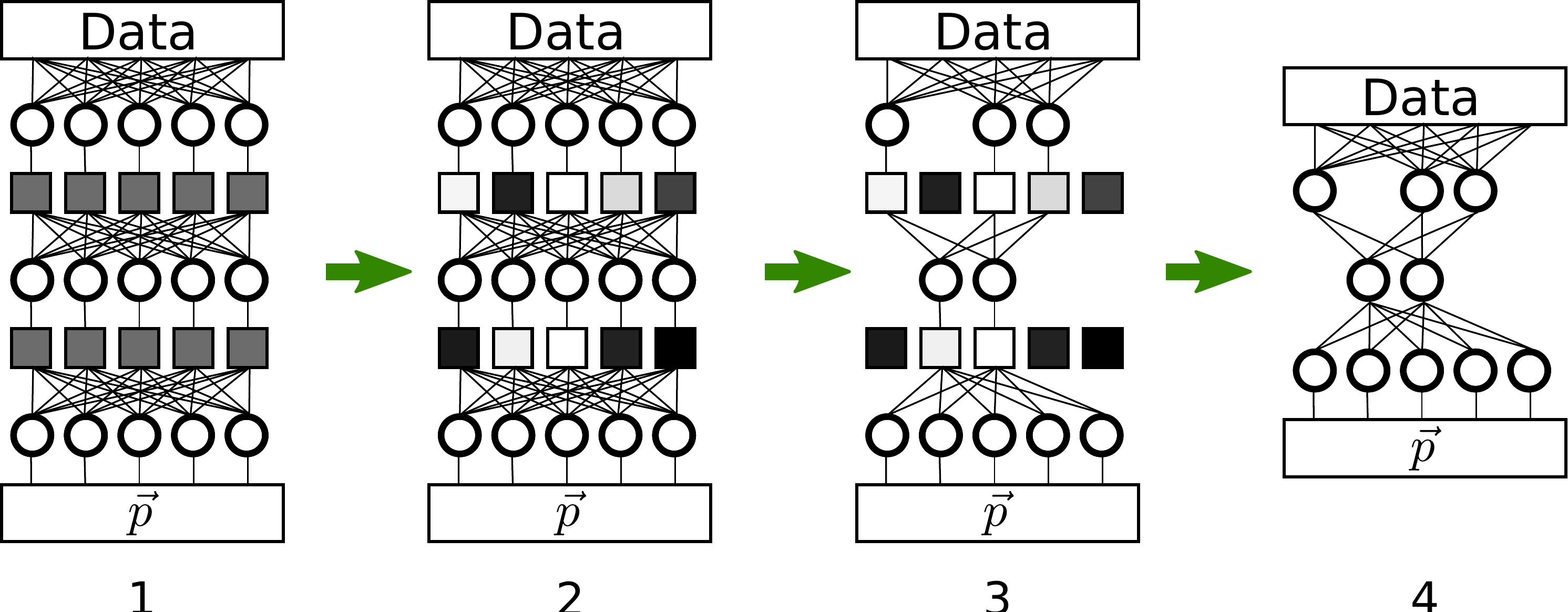}
	
	\caption{\label{fig:Neuron-pruning-algorithm} Neuron pruning algorithm graphical
		summary}
\end{figure}
By placing binary stochastic filter between NN layers it is possible
to disable the least important parts during training process. Applying
the BSF unit to every interunit connection may be used to prune
specific weights while applying it to the output of above-laying unit
will cause neuron pruning. The second approach was selected according to above-described
reasons and straightforward implementation. The algorithm is schematically
shown in the Figure \ref{fig:Neuron-pruning-algorithm}. 
\begin{enumerate}
	\item The fully-connected
	model with excess units is built, and BSF units (drawn as squares)
	are placed after each dense layer.
	\item  During training process the
	weights of neurons and BSF are optimized (values of BSF units are
	visualized as lightness of the squares, i.e. dark values correspond
	to lower weights)
	\item Weights of the filter are analyzed and neurons
	corresponding to the low weights are deleted.
	\item The BSF layers are
	removed leaving the shape-optimized and trained NN model which
	could be additionally trained for fine-tuning.
\end{enumerate}

\subsection{Kernel pruning}
Modern convolutional networks frequently contain tens of layers, having hundreds
of convolutional kernels. For example, ResNet image recognition network \cite{he2016deep}
is composed from 34 layers having up to 512 convolutional kernels. This makes such networks
extremely computationally intractable for both training and evaluation, limiting their
applications to hardware-accelerated desktops or servers, leaving aside 
highly attractive embedded and mobile systems. This issue could be solved by finding
reduced number convolution kernels, enough to maintain the classification accuracy.
This could be done with slightly modified version of BSFilter, which filters out 
feature maps, produced by convolution operation, instead of a single neurons.
Otherwise, the method is similar to the described in the previous subsection.

\subsection{Implementation}

\medskip{}

The BSF layer was implemented in Keras framework with Tensorflow back-end
\cite{chollet2015keras,tensorflow2015-whitepaper} based on above-discussed
theory. Implementation together with usage instructions is available on GitHub \footnote{\url{https://github.com/Trel725/BSFilter}}. 
All NN models used in experiments were built by dint of the same
framework.

\section{Results}

Three different datasets were used, the \emph{Wine} dataset \cite{forina2012wine},
\emph{KDD-99} \cite{Chaudhuri:2000:KFA:846183.846194} and \emph{Musk2}
dataset \cite{dietterich1997solving}. As an original KDD99 dataset
is huge (around 5 million of instances), a sample of around 1 million
instances was used instead. The datasets summary is given in the
Table \ref{tab:Classification-datasets-summary}.
\begin{table}
\footnotesize
{{\caption{\label{tab:Classification-datasets-summary} Classification datasets
summary}
}}

{{}}%
\begin{tabular}{>{\centering}p{0.08\columnwidth}>{\centering}p{0.14\columnwidth}>{\centering}p{0.08\columnwidth}>{\centering}p{0.08\columnwidth}>{\centering}p{0.4\columnwidth}}
\hline 
 & {{Instances } } & {{Classes } } & {{Features } } & {{Description}}\tabularnewline
\hline 
\hline 
{{Wine } } & {{178 } } & {{3 } } & {{13 } } & {{Chemical analysis of a wine from three different
cultivars}}\tabularnewline
\hline 
{{KDD99 } } & {{918455 } } & {{21 } } & {{41 } } & {{Wide variety of intrusions data simulated in a military
network environment. }}\tabularnewline
\hline 
{{Musk2 } } & {{6598 } } & {{2 } } & {{168 } } & {{Conformation of molecules judged to be musk or non-musk}}\tabularnewline
\hline 
\end{tabular}
\end{table}
 Datasets were standardized by removing mean and scaling to unit variance.
Musk2 and KDD99 datasets contain categorical data labels which were
encoded as integers before standardization. 
\subsection{Deep NN classifier}

NN classifier was trained on each dataset and corresponding accuracies
were saved as original. Feature selection was performed as it is described
in the Subsection \ref{subsec:Deep-NN-classifier}, and accuracy estimation
was repeated for truncated versions of datasets. Obtained data is
summarized in the Table \ref{tab:Selection-results} together with
reference accuracies, found in the literature. Additionally, the experiment
was repeated for KDD99 dataset but with 12 least important features
(corresponding BSF weights are below 30th percentile). Maximal accuracy
has achieved 57.82 \% in that case.

The implementation of silhouettes from the scikit-learn project
\cite{scikit-learn} was used in the experiment. The method was applied
to the original and truncated versions of the \emph{Wine} dataset. 
Silhouettes coefficients (\ref{silhouttes}) for these datasets were calculated as 0.2798 
and 0.3785 respectively, which is an evidence of enhanced  cluster separation
after feature selection.

\label{stability_estimating}
Stability of a binary filter was estimated by varying the structure
of NN classifier, training the system with \emph{KDD99} dataset and
analyzing the features of the highest importance (above 80th percentile). Each
experiment was repeated 4 times and selected features were summarized
in the Table \ref{tab:selection-results-stability}. The column NN
structure contains the number of units in each layer and their activation
functions. Additionally, the output of every model consists of a softmax-activated
layer, which is not shown in the table. 

\begin{table}
	\footnotesize
	{{\caption{\label{tab:Selection-results} Comparison of classification results
				for original and truncated versions of three given datasets (cross-validated)}
	}}
	
	{{}}%
	\begin{tabular}{>{\centering}p{0.35\columnwidth}>{\centering}p{0.15\columnwidth}>{\centering}p{0.15\columnwidth}>{\centering}p{0.15\columnwidth}}
		\toprule 
		& {{Wine } } & {{KDD99 } } & {{Musk2}}\tabularnewline
		\midrule
		\midrule 
		{{Training accuracy, original } } & {{1.000 } } & {{0.9993 } } & {{1.000}}\tabularnewline
		\midrule 
		{{Validation accuracy, original } } & {{0.9891 } } & {{0.9993 } } & {{0.9066}}\tabularnewline
		\midrule 
		{{Number of selected features } } & {{6 } } & {{8 } } & {{42}}\tabularnewline
		\midrule 
		{{Training accuracy, truncated } } & {{1.000 } } & {{0.9975 } } & {{1.000}}\tabularnewline
		\midrule 
		{{Validation accuracy, truncated } } & {{0.9889 } } & {{0.9975 } } & {{0.9251}}\tabularnewline
		\midrule 
		{{Reference accuracy } } & {{0.9887 \cite{misra2007functional} } } & {{0.9941 \cite{shrivas2014ensemble} } } & {{0.8920 \cite{dietterich1997solving}}}\tabularnewline
		\bottomrule
	\end{tabular}
\end{table}

\subsection{Convolutional Classifier }
\label{sec:conv_class}
The MNIST dataset from the work of Le Cun et al. \cite{lecun1990handwritten}
was used for CNN training. The original set contains 60000 handwritten
digit images of size 28x28 pixels with labels, from which 15000
were randomly sampled as training dataset and another 5000 as validation one. The implemented
model structure is shown in the appendix (Table \ref{tab:Convolutional-NN-classifier}).
The model has achieved training and validation accuracies of 1.000 and 0.990
respectively. BSF layer was added to model and training process was
repeated. The weights of the layer were visualized (Fig. \ref{fig:spectra}
(a)) and clearly correspond to the central part of the image where the majority of
 information about the digit is located. An animated demonstration of the
training process could be found in supplementary materials.

\subsubsection{Spectra recognition}

A classifier model was trained on spectra dataset, a sample of which
is shown in Fig. \ref{fig:spectra} (b). Weights of the layer
are then visualized (Fig. \ref{fig:spectra} (c)) and treated as
regions of interest of spectra. 

\subsection{Neuron pruning}

The experiment was performed with the datasets, used in the Section
\ref{subsec:Deep-NN-classifier}. The implementation
utilizes additional functions for model rebuilding \cite{whetton2018kerassurgeon}.
The model with excess neurons was trained on each dataset until convergence
and then pruned. The weights of the shape-optimized model were reset and
model was trained until convergence (number of epochs were determined experimentally to guarantee no further loss decrease) in 10-fold cross-validation manner,

Mean values for accuracy are tabulated in the Table \ref{tab:pruning_results}.

\begin{table}
\footnotesize
{{\caption{\label{tab:pruning_results} Summary of the NN pruning results for
different datasets and corresponding accuracies}
}}

{{}}%
\begin{tabular}{>{\centering}p{0.35\columnwidth}>{\centering}p{0.15\columnwidth}>{\centering}p{0.15\columnwidth}>{\centering}p{0.15\columnwidth}}
\toprule 
 & {{Wine } } & {{KDD99 } } & {{Musk2}}\tabularnewline
\midrule
\midrule 
{{Training accuracy, original } } & {{1.000 } } & {{0.9993 } } & {{1.000}}\tabularnewline
\midrule 
{{Validation accuracy, original } } & {{0.9891 } } & {{0.9993 } } & {{0.9066}}\tabularnewline
\midrule 
{{Original number of units/weights } } & {{56/953 } } & {{185/9430 } } & {{666/138778}}\tabularnewline
\midrule 
{{Number of units/weights after pruning } } & {{33/469 } } & {{31/595 } } & {{57/4103}}\tabularnewline
\midrule 
{{Training accuracy, pruned } } & {{1.000 } } & {{0.9993 } } & {{1.000}}\tabularnewline
\midrule 
{{Validation accuracy, original } } & {{0.9830 } } & {{0.9993 } } & {{0.9987}}\tabularnewline
\bottomrule
\end{tabular}

{{\medskip{}
 } }
\end{table}

\subsection{Kernel pruning}

Two datasets were used in the experiments, above-mentioned MNIST and CIFAR10, containing
60000 32x32 images, grouped into 10 categories \cite{krizhevsky2009learning}. Due to the significant
difference between datasets, two convolutional models have been used. The MNIST dataset was 
classified with the same CNN as described in section \ref{sec:conv_class}. CIFAR10 dataset was
classified with simple deep CNN, described in Keras documentation \cite{keras_cnn}.
In the first stage, both datasets were classified by training for $N_{tr}$ epochs and then fine-tuned 
by additional training for $N_{ft}$ epochs with optimizer, learning rate of which was reduced in 10
times.
In the second stage, the modified BSFilter layer was placed after each of the convolutional layers, the
model was trained by the same number of epochs, pruned and fine-tuned for $N_{ft}$ epochs.

\begin{figure}
	\centering
	\begin{minipage}{0.48\columnwidth}%
		\subfloat[]{\includegraphics{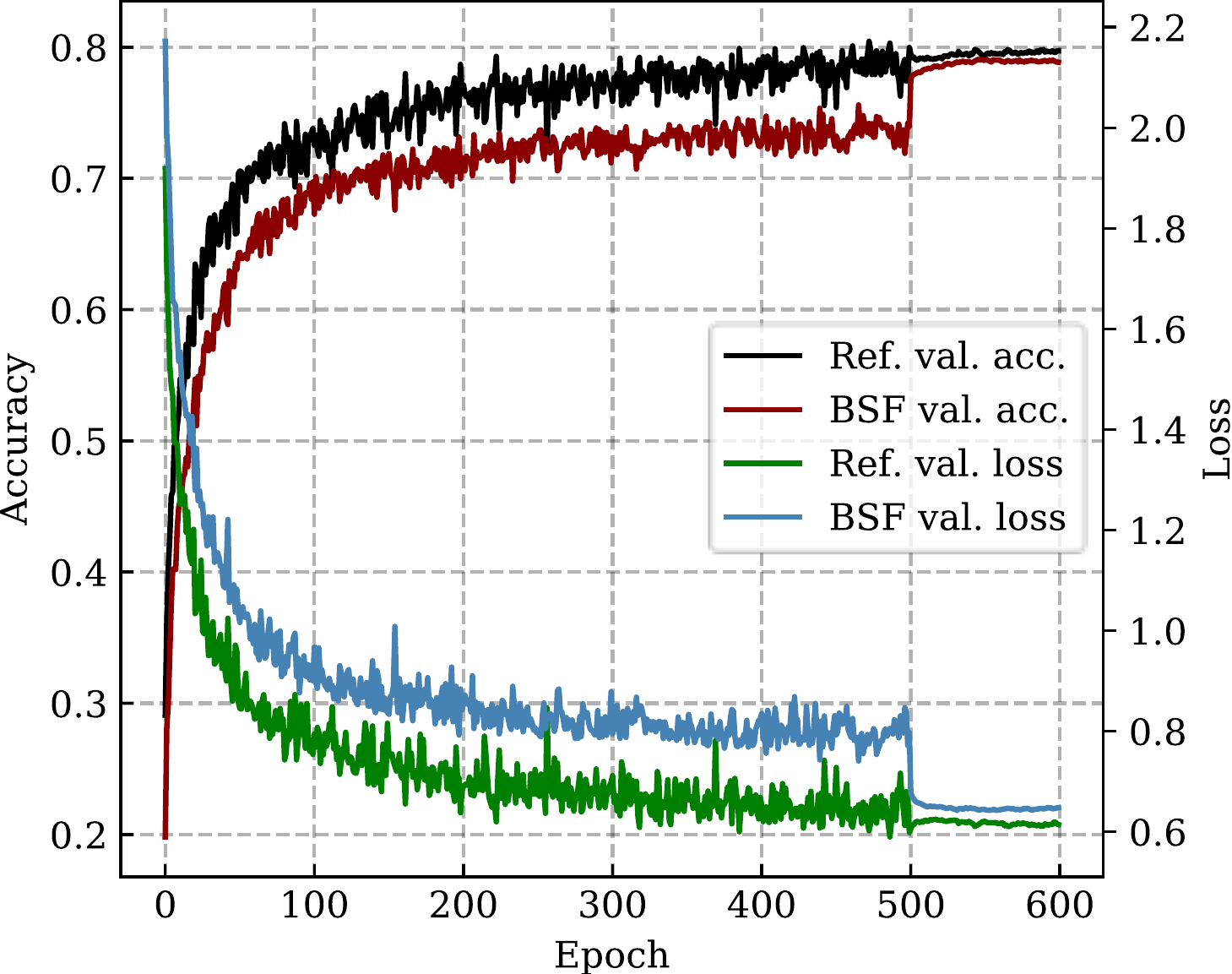}}
	\end{minipage}
	\begin{minipage}{0.48\columnwidth}%
		\subfloat[]{\includegraphics{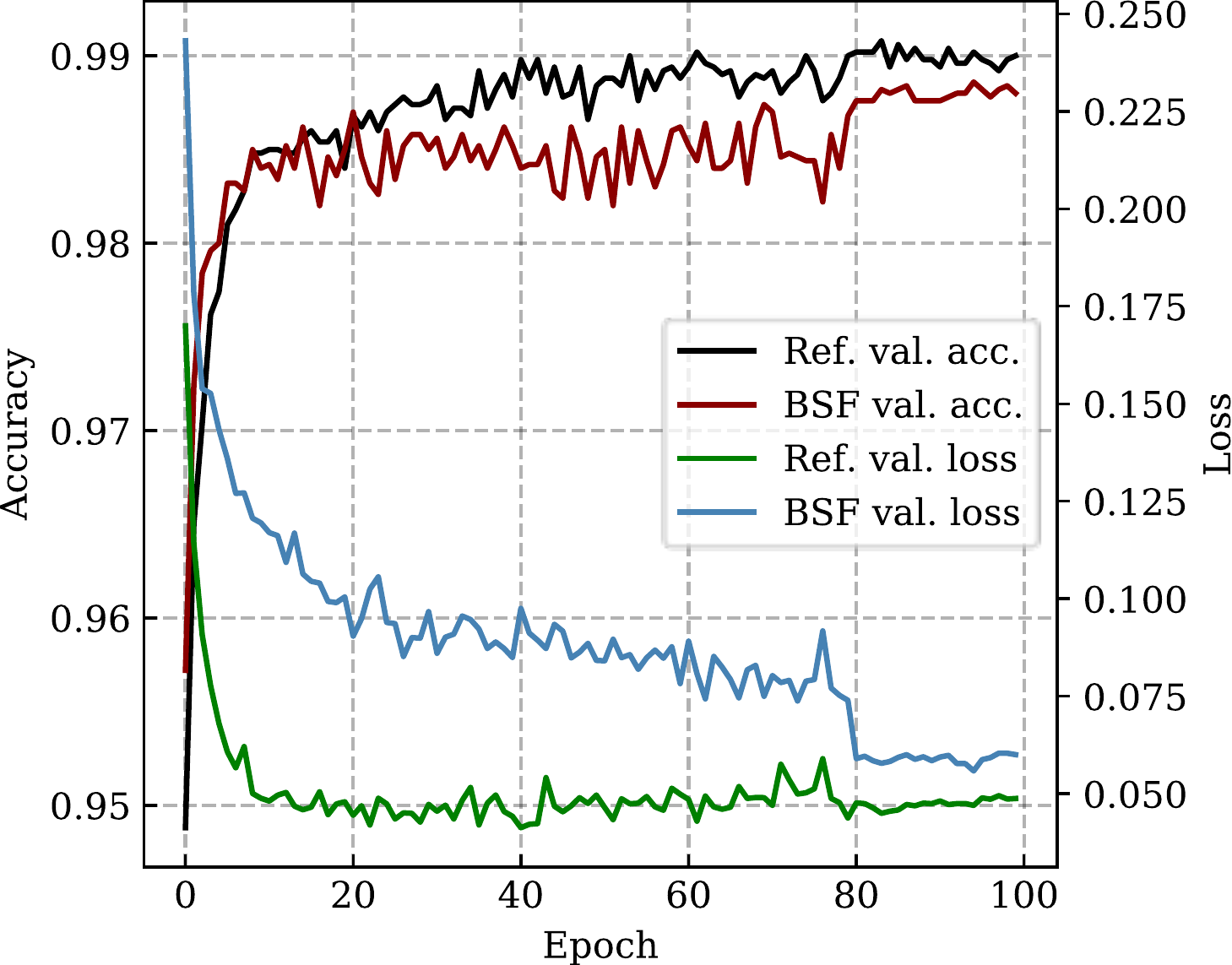}}
	\end{minipage}
	
	\caption{Validation accuracy and loss evolution during CNN models training for 
		CIFAR10 (a) and MNIST (b) datasets. Change of curve behavior at after 500$^{th}$ epoch (a)
	and after 80$^{th}$ epoch (b) is connected with change of learning rate and model pruning.}
	\label{fig:cifar10}
\end{figure}

Validation accuracy and loss evolution during training for both datasets is presented in the Figure \ref{fig:cifar10}.
Final accuracies, losses, number of weights and model size are tabulated in the Table \ref{tab:kernelpruning}.

\begin{table}
	\begin{tabular*}{1\textwidth}{@{\extracolsep{\fill}}cccccc}
		\toprule 
		Model & L1 coef. & Accuracy & Loss & Weights & Size\tabularnewline
		\midrule
		\midrule 
		MNIST & --- & 0.9900 & 0.0488 & 1,199,882 & 13.8 MiB\tabularnewline
		\midrule 
		MNIST+BSF & \num{1e-3} & 0.9880 & 0.0600 & 279,508 & 3.2 MiB\tabularnewline
		\midrule 
		MNIST+BSF & \num{1e-2} & 0.9818 & 0.1131 & 112,144 & 1.3 MiB\tabularnewline
		\midrule 
		CIFAR10 & --- & 0.7972 & 0.6143 & 1,250,858 & 9.6 MiB\tabularnewline
		\midrule 
		CIFAR10+BSF & \num{1e-4} & 0.7885 & 0.6479 & 391,519 & 3.1 MiB\tabularnewline
		\midrule 
		CIFAR10+BSF & \num{1e-3} & 0.7583 & 0.7277 & 269,273 & 2.1 MiB\tabularnewline
		\bottomrule
	\end{tabular*}

	\caption{Summary of kernel pruning results. Accuracy and loss values obtained from the validation run.}
	\label{tab:kernelpruning}
	
\end{table}

\section{Discussion}

\subsection{Deep NN classifier}

From the classification results (Table \ref{tab:Selection-results})
one can conclude that correct feature selection may lead to significant
increase of model performance without considerable loss of accuracy,
as for each of three datasets the multifold dimensionality decrease was
observed, leading to the corresponding reduction of computational complexity.
Experiment with features corresponding to the lowest BSF weights proves
that these features do not contain important information for classification.

Another important consequence is the fact that BSF allows concluding
which features in the dataset actually carry information. The importance
distribution for two datasets (\emph{Wine} and \emph{KDD99}) is shown
in the Appendix. It is possible to conclude that some of the data,
present in datasets is not meaningful in the classification problem,
i.e. does not contain information about belonging to some class, like
phenols concentration in Italian wines (Fig. \ref{fig:Feature-importances-distribution_wine})
or number of accessed files in the malware traffic (Fig. \ref{fig:Feature-importances-distribution-kdd99}).
This could lead to interesting conclusions, like independence of phenols
concentration on used cultivar. The \emph{Musk2} dataset is not easily
interpretable in a similar style, but the results could probably
be even more important, as from the meaningful features the information about
molecule biologically active sides could be extracted, which may be
the valuable information for drug designers.

Because of the high popularity of KDD99 dataset, it is possible to additionally
compare feature selection results with results of other researchers,
which are tabulated in the Table \ref{tab:Comparison-of-feature}.
In addition to 8 selected features, mentioned above, the truncated
dataset was enlarged to 14 features, which is the minimal dimensionality
leading to zero decrease of accuracy in comparison to the original dataset.
It is possible to conclude from the table, that BSF surpasses other
feature selection methods in terms of accuracy per number of features.
It is worth mentioning that in the works used for comparison, best
Validation accuracy was used as a measure of classification success, while
present work uses the cross-validated average accuracy.
\begin{table}[t]
\footnotesize
{{\caption{\label{tab:Comparison-of-feature} Comparison of feature selection
results for KDD99 dataset}
}}

{{}}%
\begin{tabular}{>{\centering}p{0.35\columnwidth}>{\centering}p{0.15\columnwidth}>{\centering}p{0.15\columnwidth}>{\centering}p{0.15\columnwidth}}
\toprule 
{{Method } } & {{Features } } & {{Accuracy } } & {{Reference}}\tabularnewline
\midrule
\midrule 
{{Fast correlation-based filtering, Naive Bayes/decision
tree } } & {{12 } } & {{0.9460 } } & {{\cite{deshmukh2015improving}}}\tabularnewline
\midrule 
{{Decision tree, simulated annealing } } & {{28 } } & {{0.9936 } } & {{\cite{lin2012intelligent}}}\tabularnewline
\midrule 
{{Support vector machine, particle swarm optimization
} } & {{26 } } & {{0.9945 } } & {{\cite{lin2008particle}}}\tabularnewline
\midrule 
{{Supported vector machine, simulated annealing } } & {{25 } } & {{0.9942 } } & {{\cite{lin2008parameter}}}\tabularnewline
\midrule 
{{Supported vector machine, decision tree, simulated
annealing } } & {{23 } } & {{0.9996 } } & {{\cite{lin2012intelligent}}}\tabularnewline
\midrule 
\emph{Binary stochastic filtering}{{
} } & \emph{8}{{ } } & \emph{0.9975}{{ } } & {{-}}\tabularnewline
\midrule 
\emph{Binary stochastic filtering}{{
} } & \emph{14}{{ } } & \emph{0.9993}{{ } } & {{-}}\tabularnewline
\bottomrule
\end{tabular}
\end{table}

\begin{figure}[t]
	\begin{minipage}{0.45\columnwidth}%
		\subfloat[]{\includegraphics{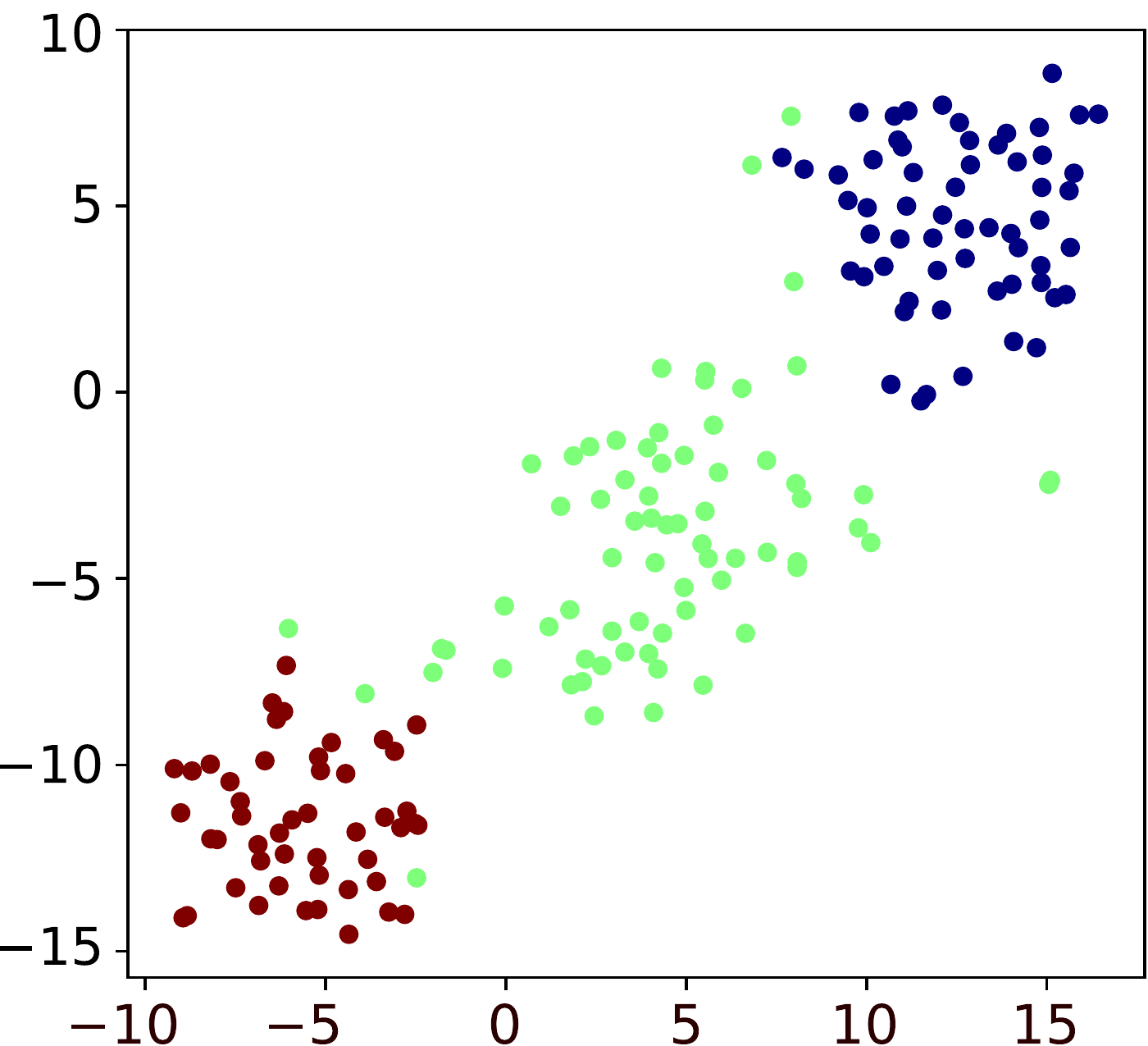}
			
		}%
	\end{minipage}%
	\begin{minipage}{0.45\columnwidth}%
		\subfloat[]{\includegraphics{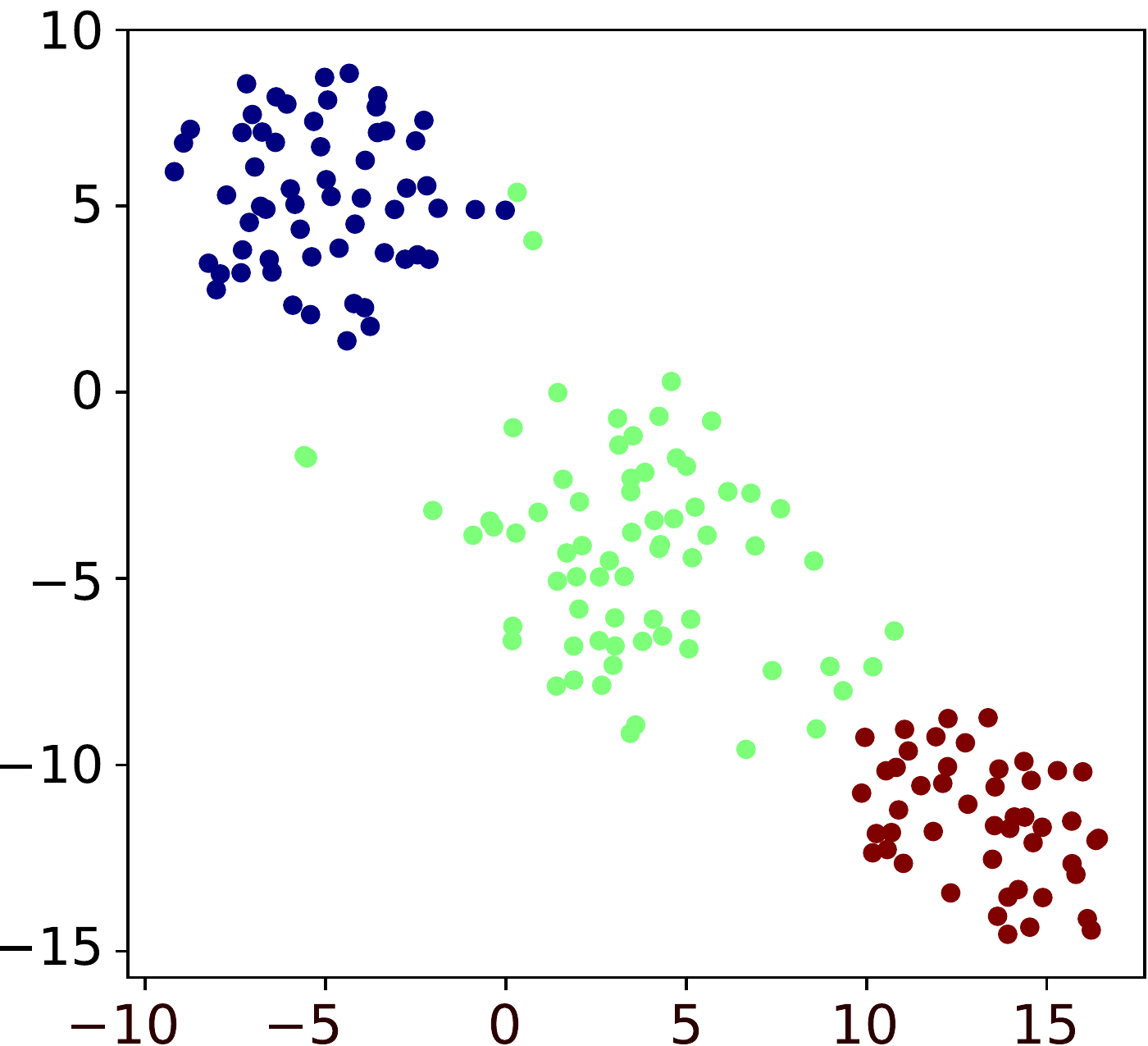}
			
		}%
	\end{minipage}
	
	\caption{\label{fig:Visual-representation-tsne} Visual representation of the (a)
		original Wine dataset and (b) its truncated version, containing 6 most
		important features. It is visible that cluster separation has
		increased after removing unimportant features. Data was compressed to 2 dimensions by the T-SNE algorithm.}
\end{figure}

Values of silhouettes coefficient for the truncated Wine dataset were
observed to be higher than for original version. Thus, removing 
unimportant features, that behave as noise, actually improves the dataset
classification. Both versions were additionally processed with the
well-known T-SNE algorithm \cite{maaten2008visualizing}, which allows high-dimensional data visualizing in the two-dimensional
plot, which makes it possible to see enhanced separation by naked eye.
Generated images are shown in the Fig. \ref{fig:Visual-representation-tsne}

The stability estimating experiments demonstrate the fact that the
BSF feature selector is relatively stable with only a small portion
of features changing in different experiments. It was found by manual
observation of selected features importances, that differences in selections
occurs from the threshold separation into important and unimportant.
During selection, weights of some features are approaching
unity, while other ones are approaching zero. Stochastic nature of selection
process causes small deviations in the weights of most important features
($w\approx1$), which leads to fluctuations in selection results. This
could be observed in the animation visualizing the BSF weights evolution
during training process available in the supplementary materials.

\subsubsection{Spectra recognition}

Good correspondence between high-weight areas, threated as regions of interest and
vibrational frequencies of the DNA constituents \cite{madzharova2016surface}
was observed, which was expected from the chemical point of view.
This method allows one to obtain information about subtle chemical
changes in difficult samples. 
\subsection{Self-optimizing neural network}

Minimal decrease of accuracy was observed from the data, given in
the Table \ref{tab:pruning_results}, while number of weights was
reduced significantly, from 2 to approximately 34 times. This corresponds
to the enhancement of performance for NN prediction of the same order,
as computational complexity is directly connected to the number of
weights. Such optimization could be especially beneficial for the embedded NN
applications having limited computational resources. Obviously, the scale of NN pruning depends on initial
redundancy but from the experiments it is possible to make empirical
remark that satisfactory starting number of neurons in the hidden layers is equal
to the dimensionality of input data. Another important outcome is
the fact that \emph{Musk2} dataset, which was the most efficiently
pruned, exhibits a considerable increase of Validation accuracy, i.e. improved
generalization in correspondence with the \cite{lecun1990optimal}.
Obtained cross-validated accuracy (0.9987) exceeds all experimental
accuracies for the given dataset, found in literature (0.8920, 0.8333,
0.9600) \cite{dietterich1997solving,zhang2002dd,zhang2004improve}.

\subsection{Kernel pruning}

\begin{figure}
	\centering
	\begin{minipage}{0.48\columnwidth}%
		\subfloat[]{\includegraphics{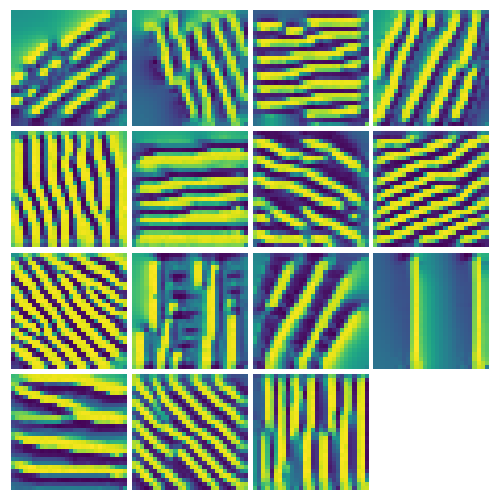}}
	\end{minipage}
	\begin{minipage}{0.48\columnwidth}%
		\subfloat[]{\includegraphics{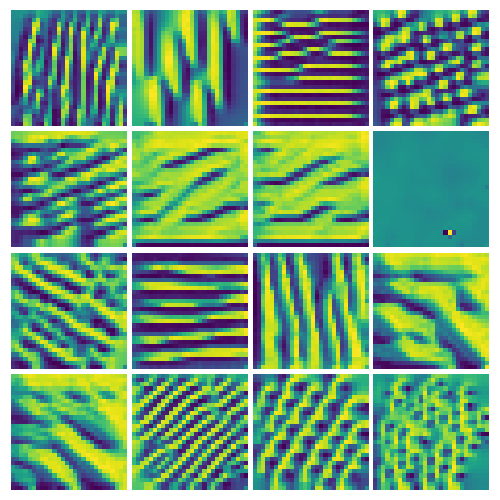}}
	\end{minipage}
	
	\caption{Images, maximizing activation of the last convolutional layer
	for pruned CNN (a)  and original CNN (b). For the latter, 16 images were randomly
	selected from the total of 64.}
	\label{fig:activationsbsf}
\end{figure}

Results, summarized in the Table \ref{tab:kernelpruning} demonstrate that 
kernel filtering with successive pruning is able to significantly
decrease computational complexity of the image-recognition CNN at the price
of minor accuracy drop. Precisely, more than threefold decrease in number of weights
for the CIFAR10 dataset was achieved, accompanied by a change in accuracy by less than 1 \%.
Stronger penalization of the BSF layer lead to even more significant pruning, leaving around 20 \%
from the initial network weights. Nevertheless, in that case the decrease in accuracy 
was around 4 \%, which is unacceptable. This observation may be used as an indicator of
the optimal number of weights in the network.

During experiments with MNIST dataset, the number of weights was reduced in
4.3 times with change of accuracy in 0.2 \%.
In the experiments with increased penalization coefficient more than tenfold 
reduce in weights was reached at the cost of accuracy decrease by 0.82 \%.

From the results above follows the fact that efficiency of kernel pruning
(same as neuron pruning) strongly depends on the initial redundancy of neural network.
By tuning penalization coefficient and observing the change of accuracy and loss it is possible
to find the optimal balance between size of the CNN and its predictive ability. 

Other interesting conclusions could be done from the analysis of the convolutional 
kernels, optimized during CNN training. To visualize convolutions the input images,
maximizing activation of a given kernel were generated with keras-vis library \cite{raghakotkerasvis}.
This method was applied to the last convolutional layer of the MNIST classifier, having 64 kernels
in the original model and 15 in the pruned version. Results, presented in the Figure \ref{fig:activationsbsf}
demonstrates that pruned model is prone to learn smaller number of simpler kernels, mostly differently
inclined lines, which encode majority of the information about the number. Unconstrained original model
learns more complex patterns, such as curves, crosses and others which are difficult to interpret by a human,
representing less informative details of the image. This gives a clue to the question of how the majority of
kernels could be removed from the model almost without drop in generalization: the CNN is focused on the most
informative features, ignoring the details.

\section{Conclusion}

Binary stochastic filter is a straightforward method for estimating
importances of input features or specific units. In comparison
to existing works, which select features by repetitive evaluating performance of subset,
BSF works \textit{in situ} with the whole dataset, minimizing feature involvement in the training
process.
The method has a wide variety of possible applications, including feature
selection and NN size minimization. Thank its layered structure
it can be easily introduced to different kinds of NN without any additional overhead.
During experiments, BSF selected 8 features
from 42 in the KDD99 dataset, bearing enough information for close-to-one prediction accuracy. 
Such kind of feature selection is of interest
in natural sciences, as with its help one could make conclusions about
processes, generating information, contained in dataset. Suggested
neuron pruning algorithm has led to a significant increase of NN performance
in experiments, decreasing the number of weights in $\approx$34 times
for Musk2 dataset. What even more substantial, after shape optimization 10-fold
crossvalidation test demonstrated considerable increase of validation
accuracy, proving better generalization. Kernel pruning, implemented
with BSF has achieved more than tenfold decrease of model size, which
makes it a promising method for transferring NN to the mobile and embedded systems.

Thus, BSF could become a useful
tool in the different fields of machine learning. Further work will include
investigation of the weights pruning, i.e. network sparsifying,
generalizing of BSF layer for filtering convolution maps, i.e. kernel sparsifying
and application
of the method to more complex NN structures, such as adversarial networks
and autoencoders. 

\bibliographystyle{ieeetr}
\bibliography{./bsfilter}

\pagebreak{}

\appendix

\section{Proof of gradients independence on introducing BSF layer}

\begin{figure}[H]
	\begin{minipage}[t]{0.43\columnwidth}%
		\subfloat[]{\includegraphics{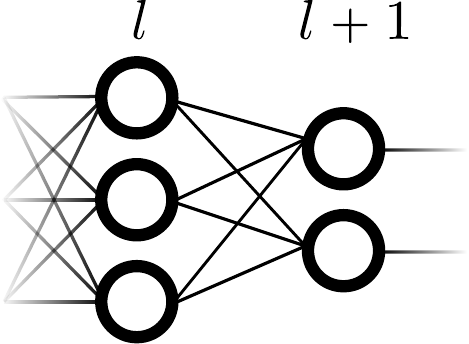} 
			
		}%
	\end{minipage}%
	\begin{minipage}[t]{0.5\columnwidth}%
		\subfloat[]{\includegraphics{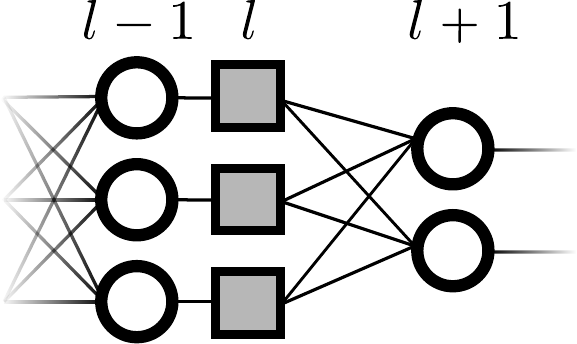}
			
		}%
	\end{minipage}
	
	\caption{\label{fig:illustration_prrof}}
	
\end{figure}

\label{proof:gradients_independence}
\begin{proof}
	
	Suppose the multilayer perceptron is given. Let us focus on
	the two hidden layers in the perceptron, having indices $l$ and $l+1$
	and containing $i$ and $j$ neurons in each respectively (Fig. \ref{fig:illustration_prrof}
	(a)). Suppose the partial derivatives of loss function $L$ with respect
	to the neuron pre-activation $z_{j}^{l+1}$ is given. Let us call
	this value $\delta_{j}^{l+1}=\frac{\partial L}{\partial z_{j}^{l+1}}$
	an error of the $j$ neuron in the $l+1$ layer, which could be computed
	from the classical backpropagation algorithm, and is enough for the
	computing of derivatives with respect to weights, as $\frac{\partial L}{\partial w_{ij}^{l-1}}=\frac{\partial L}{\partial z_{j}^{l+1}}\frac{\partial z_{j}^{l+1}}{\partial w_{ij}^{l+1}}$.
	Now the error of the neuron in $l$ layer could be calculated using
	chain rule as
	\begin{equation}
	\text{\ensuremath{\delta_{i}^{l}=}}\sum_{j}\frac{\partial L}
	{\partial z_{j}^{l+1}}\frac{\partial z_{j}^{l+1}}
	{\partial a_{i}^{l}}\frac{\partial a_{i}^{l}}
	{\partial z_{i}^{l}}\label{eq:error_of_neuron}
	\end{equation}
	where $a_{i}^{l}$ stands for activation (output) of the $i$ neuron
	in $l$ layer. $\frac{dz_{j}^{l+1}}{\partial a_{i}^{l}}$ could be
	derived from the equation describing neuron, $z_{j}^{l+1}=\sum_{i}a_{i}^{l}w_{ij}^{l+1}+b_{j}^{l+1}$
	($w_{ij}^{l+1}$ corresponds to weight, connecting the $i^{th}$neuron
	from $l$ with $j^{th}$ neuron in $l+1$, $b_{j}^{l+1}$ to corresponding
	neuron bias). As activation $a_{i}^{l}$ is independent on other neurons
	in $l$ layer, the sum could be omitted. By differentiating: 
	\begin{equation}
	\frac{\partial z_{j}^{l+1}}{\partial a_{i}^{l}}=w_{ij}^{l+1}\label{eq:dz/da}
	\end{equation}
	Last term $\frac{\partial a_{i}^{l}}{dz_{i}^{l}}$ could be derived
	at a similar manner, as $a_{i}^{l}=act(z_{i}^{l})$, where $act(z)$
	is activation function, 
	\begin{equation}
	\frac{\partial a_{i}^{l}}{\partial z_{i}^{l}}=act'(z_{i}^{l})\label{eq:da/dz}
	\end{equation}
	Substituting \ref{eq:dz/da} and \ref{eq:da/dz} into \ref{eq:error_of_neuron},
	the equation for computing error of the neurons in $l$ layer could
	be obtained ($act'(z_{i}^{l})$ is independent on indices of neurons
	in $l+1$ layer, so it could be factored out):
	
	\begin{equation}
	\text{\ensuremath{\delta_{i}^{l}=}}act'(z_{i}^{l})\sum_{j}\delta_{j}^{l+1}w_{ij}^{l+1}\label{eq:error_mlp}
	\end{equation}
	Now suppose a BSF layer is introduced in between two classical layers,
	as it shown in the Fig. \ref{fig:illustration_prrof} (b). Now numbering
	will be performed by indices $i$, $j$, $k$ for every consequent
	layer. Then 
	\[
	\text{\ensuremath{\delta_{i}^{l-1}=}}\sum_{k}\frac{\partial L}{\partial z_{k}^{l+1}}\frac{\partial z_{k}^{l+1}}{\partial a_{j}^{l}}\frac{\partial a_{j}^{l}}{\partial z_{j}^{l}}\frac{\partial z_{j}^{l}}{\partial a_{i}^{l-1}}\text{\ensuremath{\frac{\partial a_{i}^{l-1}}{\partial z_{i}^{l-1}}}}
	\]
	Two members of this formula are known from previous derivation, $\frac{\partial a_{j}^{l}}{\partial z_{j}^{l}}$
	which is a derivative of BSF unit with respect to its input is equal
	to 1 from the definition of ST estimator. $\frac{\partial z_{j}^{l}}{\partial a_{i}^{l-1}}=1$,
	as BSF unit has only one input, thus $z_{j}^{l}=a_{i}^{l-1}$. Last
	term is similar to \ref{eq:da/dz}. Substituting everything, factoring
	out independent on sum member, obtaining:
	
	\[
	\text{\ensuremath{\delta_{i}^{l-1}=}}\ensuremath{act'(z_{i}^{l-1})}\sum_{k}\delta_{k}^{l+1}w_{jk}^{l+1}
	\]
	By comparing this equation with \ref{eq:error_mlp}, one can conclude
	that BSF layer actually does not influence the backpropagation process
	by other way than filtering out some interneuron connections.
	This means that under $l1$ regularization constraints of BSF layer, NN 
	will tend to find some minimal number of neurons, corresponding to the
	smallest loss function.
	
\end{proof}

\section{}

Details about convolutional NN, used in the MNIST images classification (\ref{image_recognition_section}) are given in the table \ref{tab:Convolutional-NN-classifier}.

\begin{table}
\footnotesize
{{\caption{\label{tab:Convolutional-NN-classifier} Architecture of the convolutional
NN classifier. Numerical parameters for convolution, max pooling and
fully connected layers correspond to kernel size, pool size and number
of neurons respectively.}
}}%
\begin{tabular}{>{\raggedright}p{0.3\columnwidth}>{\raggedright}p{0.3\columnwidth}>{\raggedright}p{0.3\columnwidth}}
\toprule 
{Layer} & {Parameters} & {Activation}\tabularnewline
\midrule
\midrule 
{Convolution} & {3x3, 32 kernels} & {relu}\tabularnewline
\midrule 
{Convolution} & {3x3, 64 kernels} & {relu}\tabularnewline
\midrule 
{Max pooling} & {2x2} & ---\tabularnewline
\midrule
{Flatten} & {---} & ---\tabularnewline
\midrule 
{Fully connected} & {128} & {relu}\tabularnewline
\midrule 
{Dropout} & {p=0.5} & {---}\tabularnewline
\midrule 
{Fully connected} & {10} & {softmax}\tabularnewline
\bottomrule
\end{tabular}
\end{table}

\begin{figure}
\includegraphics{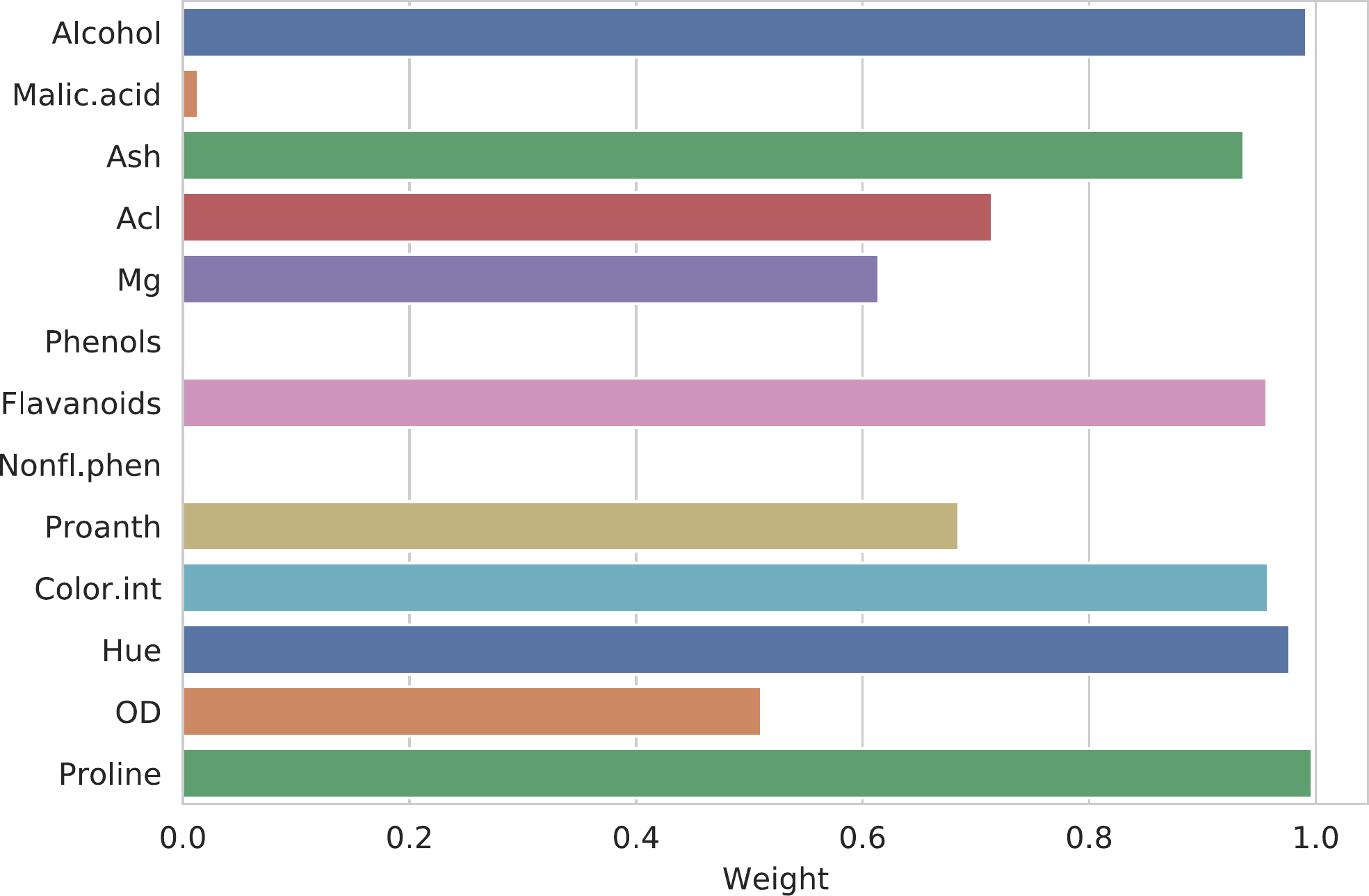}

\caption{\label{fig:Feature-importances-distribution_wine}Feature importances
distribution for the \emph{Wine} dataset}
\end{figure}

\begin{figure}
\includegraphics{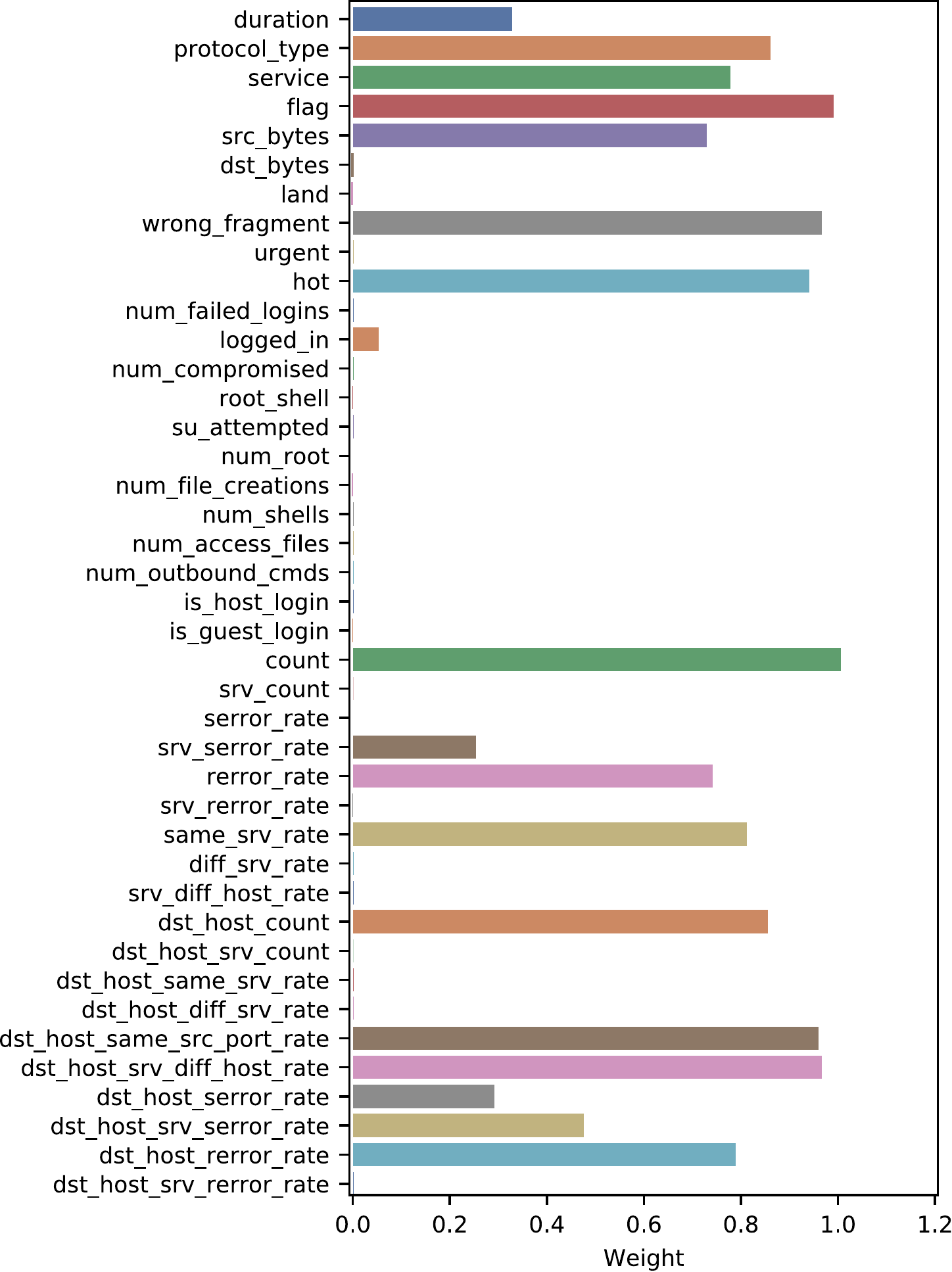}

\caption{\label{fig:Feature-importances-distribution-kdd99}Feature importances
distribution for KDD99 dataset}
\end{figure}

Highest importance features, selected by BSF method with different structure (\ref{stability_estimating})
of used NN is tabulated in the  Table \ref{tab:selection-results-stability}.
The diagrams demonstrating specific feature importances are given in the Figs. \ref{fig:Feature-importances-distribution_wine}
and \ref{fig:Feature-importances-distribution-kdd99}

\begin{table}
\footnotesize
{{\caption{\label{tab:selection-results-stability} Indices of selected features
for different classifier structure}
}}

\begin{tabular}
{>{\centering}p{0.25\columnwidth}>{\centering}p{0.5\columnwidth}}
\toprule 
{{NN structure } } & {{Selected features}}\tabularnewline
\midrule
\midrule 
{{41, relu}}

{{82, relu}}

{{41, relu } } & {{{[} 1, 2, 9, 22, 28, 34, 35, 36{]}}}

{{{[} 1, 3, 7, 9, 22, 34, 35, 36{]}}}

{{{[} 3, 7, 9, 22, 28, 34, 35, 36{]}}}

{{{[} 1, 7, 9, 22, 31, 34, 35, 36{]}}}\tabularnewline
\midrule 
{{41, tanh}}

{{41, relu}}

{{41, relu}}

{{41, relu } } & {{{[} 3, 7, 9, 28, 31, 34, 35, 36{]}}}

{{{[} 7, 9, 22, 28, 31, 34, 35, 36{]}}}

{{{[} 1, 7, 9, 22, 28, 31, 34, 35{]}}}

{{{[} 7, 9, 22, 28, 31, 34, 35, 36{]}}}\tabularnewline
\midrule 
{{41, tanh}}

{{82, tanh}}

{{41, tanh } } & {{{[} 1, 7, 22, 28, 31, 34, 35, 36{]}}}

{{{[} 3, 7, 9, 22, 31, 34, 35, 36{]}}}

{{{[} 3, 7, 9, 22, 28, 34, 35, 36{]}}}

{{{[} 1, 7, 22, 28, 31, 34, 35, 36{]}}}\tabularnewline
\midrule 
{{41, relu}}

{{20, relu}}

{{13, relu}}

{{10, relu}}

{{8, relu } } & {{{[} 1, 7, 9, 22, 28, 34, 35, 36{]}}}

{{{[} 3, 7, 9, 22, 28, 34, 35, 36{]}}}

{{{[} 3, 7, 9, 22, 31, 34, 35, 36{]}}}

{{{[} 1, 7, 9, 22, 28, 34, 35, 36{]} }}\tabularnewline
\bottomrule
\end{tabular}
\end{table}


\end{document}


\begin{center}\Large Supplementary materials to the work:\end{center}
		{\let\newpage\relax\maketitle}

\section*{Visualization method}
To obtain human-readable clear information about training process following procedure was 
performed. The weights of the BSF layer of the network model were saved every $k^{th}$ epoch,
where $k$ was selected to be 1 or 2 depending on the convergence rate. Every saved weights vector
was plotted as a bar plot (in case of 1D BSF layer for data classification problem) or as 
heatmap (in case of 2D BSF layer in  image recognition problem). Obtained images were used
to form GIF animation, which thus demonstrates the changes of BSF weights during training.
Detailed explanation of specific files is given below.

\subsection*{KDD99}

Ancillary files folder contains three GIF animations, visualizing the process of feature importance changing during model training with three different $l1$ regularization coefficients, equal to  0.01, 0.005 and 0.001. Those coefficient are included in the file names. It is visible that increase of regularization coefficient lead to rapid suppress of the features importances, nullifying majority of them.

\subsection*{MNIST}

A single GIF animation, which visualizes the evolution of BSF layer weights
during training of convolutional classifier on a MNIST dataset. The filtering quickly highlights the central
information-reach part, while peripheral pixels of the images are classified as unimportant.